\documentclass[journal,twoside,web]{ieeecolor}
\usepackage{tmi}
\usepackage{cite}
\usepackage{textcomp}
\usepackage{amsmath}
\usepackage{url}
\usepackage{xcolor}
\usepackage{times}
\usepackage{epsfig}
\usepackage{graphicx}
\usepackage{caption}
\usepackage{float}
\usepackage{amssymb} 
\usepackage{footmisc}
\usepackage{lineno}
\usepackage{color}
\usepackage{subfigure}
\usepackage{multirow}
\usepackage{dsfont}
\usepackage{mathtools}
\usepackage{setspace}
\usepackage{csquotes}
\usepackage{breqn}
\usepackage{lettrine}
\usepackage{upquote}
\usepackage{adjustbox}
\usepackage{wrapfig}
\usepackage{lipsum}
\usepackage{mathtools}
\usepackage{comment}
\usepackage{bm}
\usepackage{array}
\usepackage{tabu}
\usepackage{booktabs}
\usepackage{enumerate}
\DeclareCaptionFormat{myformat}{\fontsize{8}{10}\selectfont#1#2#3}
\DeclareMathAlphabet\mathbfcal{OMS}{cmsy}{b}{n}

\DeclareMathAlphabet{\pazocal}{OMS}{zplm}{m}{n}
\DeclareMathAlphabet{\mathpzc}{OT1}{pzc}{m}{it}

\usepackage[ruled,vlined]{algorithm2e}
\usepackage[colorlinks=true,linkcolor=blue]{hyperref}
\def\arrvline{\hfil\kern\arraycolsep\vline\kern-\arraycolsep\hfilneg}

\def\BibTeX{{\rm B\kern-.05em{\sc i\kern-.025em b}\kern-.08em
    T\kern-.1667em\lower.7ex\hbox{E}\kern-.125emX}}
\begin{document}
\title{Unsupervised Mutual Transformer Learning for
Multi-Gigapixel Whole Slide Image Classification}
\author{Sajid Javed, Arif Mahmood, Talha Qaiser, Naoufel Werghi, and Nasir Rajpoot, \IEEEmembership{Member, IEEE}
\thanks{S. Javed and N. Werghi are with the department of electrical engineering and computer science, Khalifa university of science and Technology, Abu Dhabi, UAE (e-mail: sajid.javed@ku.ac.ae).}
\thanks{A. Mahmood is with the department of computer science, ITU, Lahore, Pakistan).}
\thanks{T. Qaiser and N. Rajpoot is with the department of computer science, the university of Warwick, U.K.}}

\maketitle

\begin{abstract}
Classification of gigapixel Whole Slide Images (WSIs) is an important prediction task in the emerging area of computational pathology.
There has been a surge of research in deep learning models for WSI classification with clinical applications such as cancer detection or prediction of molecular mutations from WSIs. Most methods require expensive and labor-intensive manual annotations by expert pathologists. 
Weakly supervised Multiple Instance Learning (MIL) methods have recently demonstrated excellent performance; however, they still require large slide-level labeled training datasets that need a careful inspection of each slide by an expert pathologist. 
In this work, we propose a fully unsupervised WSI classification algorithm based on mutual transformer learning.
Instances from gigapixel WSI (i.e., image patches) are transformed into a latent space and then inverse-transformed to the original space. 
Using the transformation loss, pseudo-labels are generated and cleaned using a transformer label-cleaner.
The proposed transformer-based pseudo-label generation and cleaning modules mutually train each other iteratively in an unsupervised manner.
A discriminative learning mechanism is introduced to improve normal versus cancerous instance labeling. 
In addition to unsupervised classification, we demonstrate the effectiveness of the proposed framework for weak supervision for cancer subtype classification as downstream analysis. Extensive experiments on four publicly available datasets show excellent performance compared to the state-of-the-art methods. 
We intend to make the source code of our algorithm publicly available soon.
\end{abstract}


\section{Introduction}
\label{sec:introduction}
\IEEEPARstart{V}{isual} Despite significant improvements in cancer diagnosis and treatment, it remains a leading cause of death around the world \cite{fitzgerald2022future,he2019practical}, with
nearly 20 million new cancer cases yearly significantly burden the healthcare system \cite{sung2021global}.
Visual examination of tissue slides, often stained with Hematoxylin and Eosin (H\&E) dyes, has been considered the {\em gold standard} for cancer diagnosis in clinical practice \cite{rindi2018common, srinidhi2021deep, lu2021ai, lipkova2022deep}.
Modern-day digital slide scanners can digitize tissue slides into high-resolution multi-gigapixel Whole-Slide Images (WSIs) at 250{\em nm} per pixel, with each image containing several billions of pixels and making the direct applications of machine learning methods a challenge \cite{srinidhi2021deep, jaume2021quantifying, li2021dual, guan2022node, wu2022cross, zhang2022dtfd, chen2021multimodal, di2022generating, chen2022scaling}.
Computational pathology has recently emerged as an essential area that deals with the research and development of novel machine learning for gigapixel WSIs with applications to early cancer detection \cite{esteva2019guide, ardila2019end} and personalized medicine \cite{cui2021artificial, fuchs2011computational, tizhoosh2018artificial, srinidhi2021deep}.
Recent developments in the area have demonstrated excellent performance in various clinical tasks for analyzing tumor micro-environment, survival prediction, and response to therapy \cite{bilal2021development, chen2022pan, lu2021ai, lipkova2022deep, lu2021data, chen2020pathomic}.

\begin{figure}[t!]
\centering
\includegraphics[width=\linewidth]{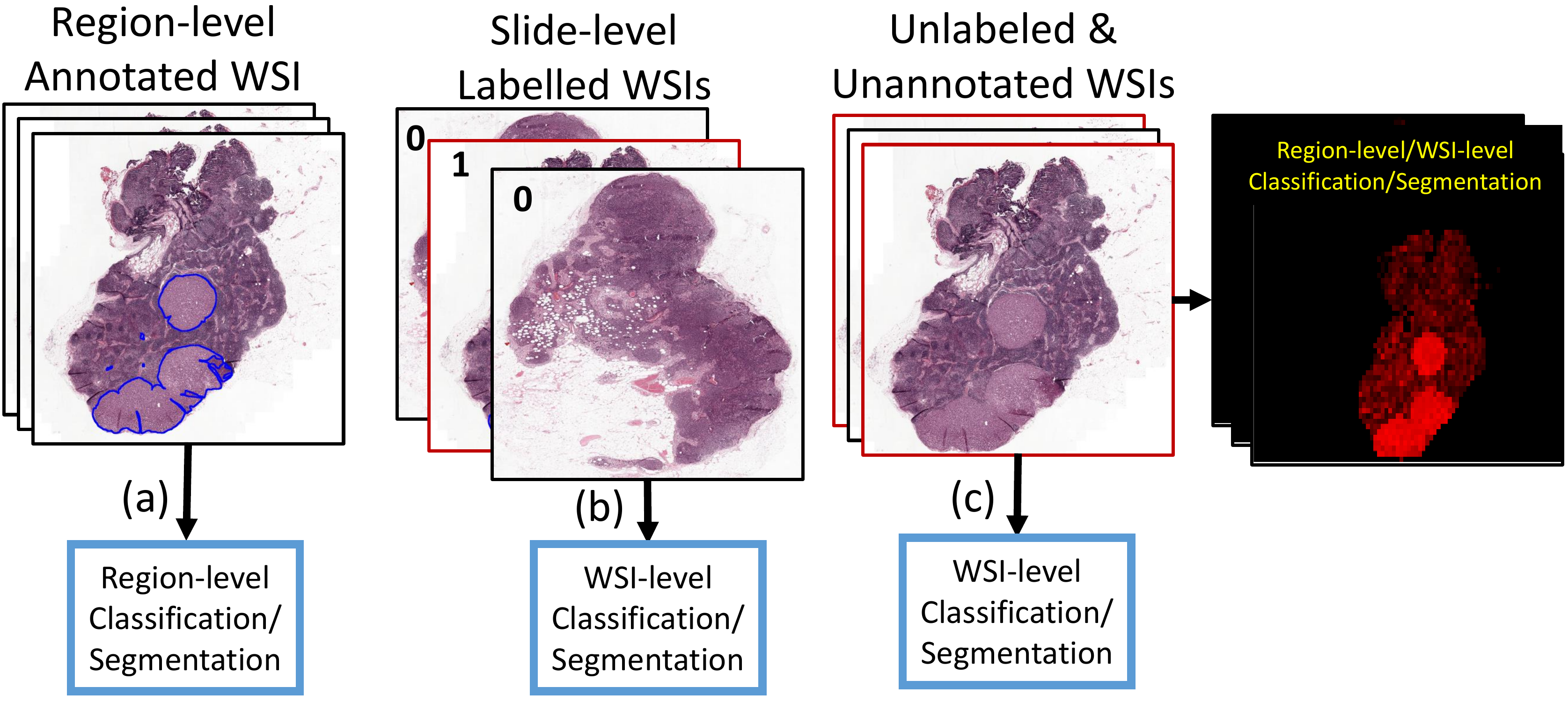}
\caption{Comparison of different types of supervision for WSI classification:
(a) Fully-supervised training requires region-level normal/tumor annotation \cite{zhang2022gigapixel,srinidhi2021deep}. 
(b) Weakly-supervised training requires slide-level labels \cite{lu2021data, li2021dual, guan2022node}.
(c) The proposed unsupervised training requires neither region-level annotations nor slide-level labels for WSI classification.
The red region in the detection maps shows the predicted tumor regions.}
\label{fig1}
\vspace{-1.5\baselineskip}
\end{figure}

Due to their huge size, annotating WSIs at the region level for fully supervised training (Fig. \ref{fig1} (a)) is a costly and time-consuming task for pathologists.
To address this challenge, Multiple Instance Learning (MIL) based weakly-supervised methods have recently been proposed that require only WSI-level labels (Fig. \ref{fig1} (b)) for WSI classification \cite{li2021dual, guan2022node, di2022generating, srinidhi2021deep, lu2021data}.
%
Although MIL methods have reduced the cost compared to the region-level annotation, an expert pathologist still has to exhaustively inspect all regions consisting of several hundreds of thousands of cells within each WSI and assign a label to each slide \cite{chen2021annotation, chuang2021identification, bejnordi2017diagnostic, tomczak2015review}.
Such inspection is still expensive and time-consuming and may limit the size of labeled WSIs dataset.
It may result in overfitting of MIL methods resulting in poorly learned features and degraded performance.
In the current work, we move one step forward by proposing a fully unsupervised WSI classification algorithm that requires unlabeled WSIs as input and learns to predict instance-level disease positive/negative predictions (Fig. \ref{fig1} (c)).
This problem is challenging yet rewarding as it may completely eradicate the cost of obtaining laborious region-level annotations and slide-level labels from pathologists and enables classification systems to be deployed without human intervention.

Unsupervised learning methods have often been considered not using any human supervision, such as different clustering methods including K-means, TSNE, and spectral clustering \cite{ezugwu2022comprehensive}.
A closely related set of methods include self-supervised learning techniques which aim to produce robust representation invariant to data augmentation  \cite{jing2020self, liu2022graph}. Such features exhibited robustness against different types of noises. Along the same line, 
Wang \textit{et al.} coupled contrastive learning with transformer models to improve the performance of self-supervised learning for WSI classification task \cite{wang2022transformer}.
Vu \textit{et al.} proposed H2T representation which are learned from unsupervised clustering techniques applied to histological image patches \cite{vu2023handcrafted}.
Chen \textit{et al.} recently proposed the HIPT by leveraging the natural hierarchical structure in WSI using self-supervised learning \cite{chen2022scaling}.
These  approaches provide robust representations, which are then utilized for WSI classification.

\begin{figure}[t!]
\centering
\includegraphics[width=\linewidth]{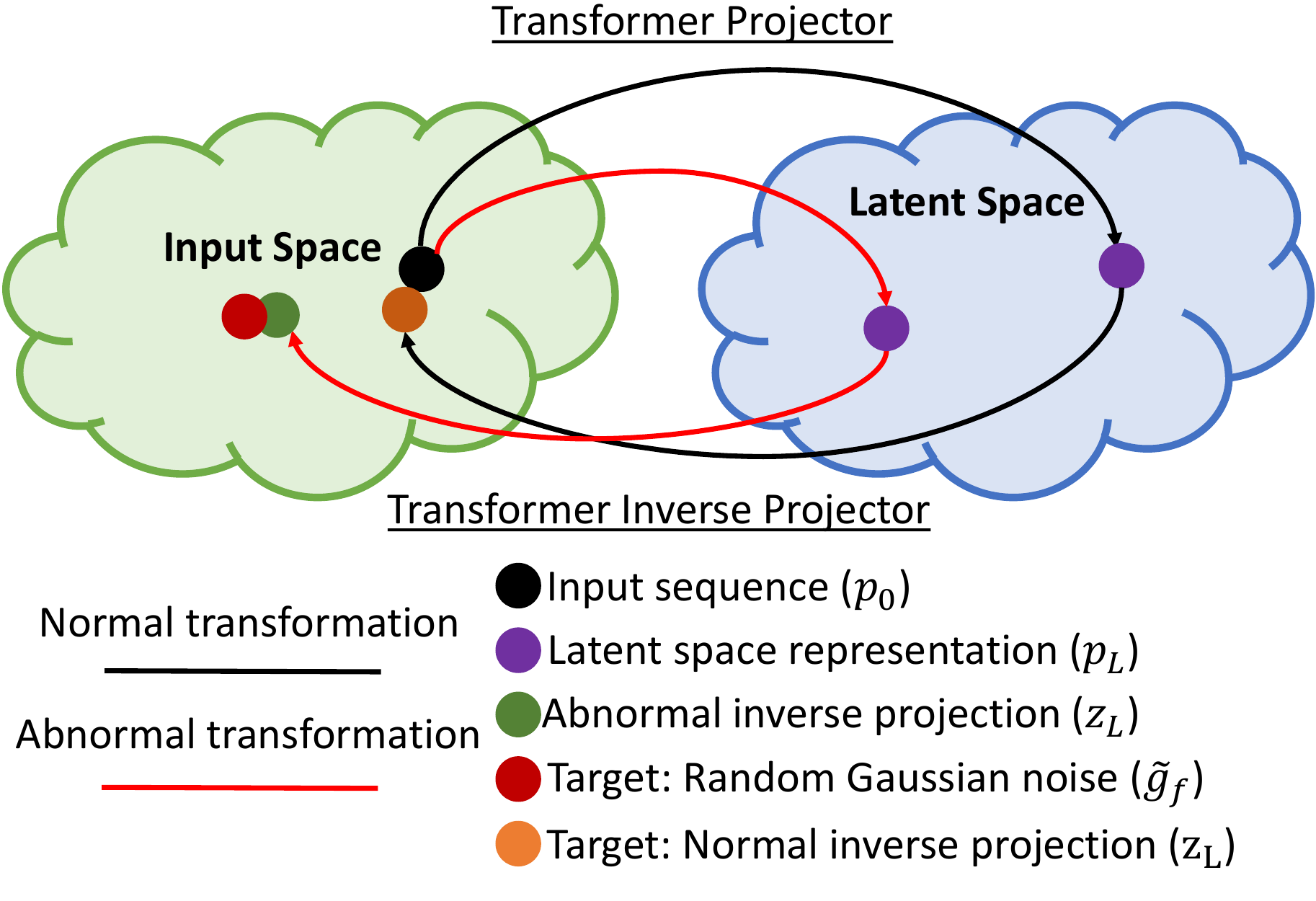}
\caption{ A latent space is learned by transformer pseudo-label generator.
The transformation error for normal instances is ensured to be low, while for tumor instances, the error is aimed to be high using discriminative learning.}
\label{fig2}
\vspace{-1.3\baselineskip}
\end{figure}

In the current work, we propose an unsupervised WSI classification algorithm that can generate slide-level labels without human intervention.
We exploit the fact that the number of disease-negative instances (WSI patches) is significantly larger than the number of disease-positive instances within WSI training datasets.
For instance, in the CAMELYON-16 dataset \cite{bejnordi2017diagnostic}, there are  0.85M positive patches and 1.38M negative patches. 
Therefore, if a learning mechanism such as autoencoder is trained without using positive or negative labels, it will better learn to represent the negative patches.
Our algorithm is inspired by observing the behavior of the autoencoder reconstruction error for the negative and the positive patches in the WSIs. We found that this error is often more significant for the positive patches when compared with their negative counterparts. 
Our interpretation is that  negative patches are more homogeneous than positive ones, which exhibit larger variation in terms of texture and patterns \cite{turashvili2017tumor, alizadeh2015toward, marusyk2010tumor}. Even though the negative patches have different categories, which  significantly differ from each other, these categories remain more homogeneous compared to the patches in the diseases-positive category.
We verify this disparity by computing the entropy of the frequency response of positive and negative patches in the CAMELYON-16 dataset. We found that the average within-patch entropy of the DCT transform of all positive patches in the CAMELYON-16 dataset is 0.881 compared to 0.556 for the negative instances. The hypothesis of positive patches being more heterogeneous than negative patches is also verified by measuring the similarity between small local windows within each patch. Using the Pearson Correlation Coefficient (PCC), we found the average within-patch PCC to be 0.357 among local windows of the positive instances as compared to the average PCC of 0.771 for negative patches.

Based on the above observations, we advocate that the reconstruction error can be leveraged to discriminate  between the positive and negative patches.
To that end,  we proposed investigating this hypothesis using a  transformer-based architecture. In the proposed algorithm, we transform input features to a latent space and then inverse transform to the original space, as shown in Fig. \ref{fig2}. 
The latent space is learned such that the transformation error is low for the disease-negative instances and high for the disease-positive ones, acting thus as an indicator of the patch type (i.e. positive or negative). 
Furthermore, we enhance the  discrimination between positive and negative patches using a  discriminative learning mechanism. Here, after the first initial iteration, the reconstruction target 
is replaced with a Gaussian random noise matrix for the large reconstruction error patches in the subsequent iterations.
We found that this arrangement improves the discrimination between the transformation of the positive and the negative instances.

\begin{figure*}[t!]
\centering
\includegraphics[width=\linewidth]{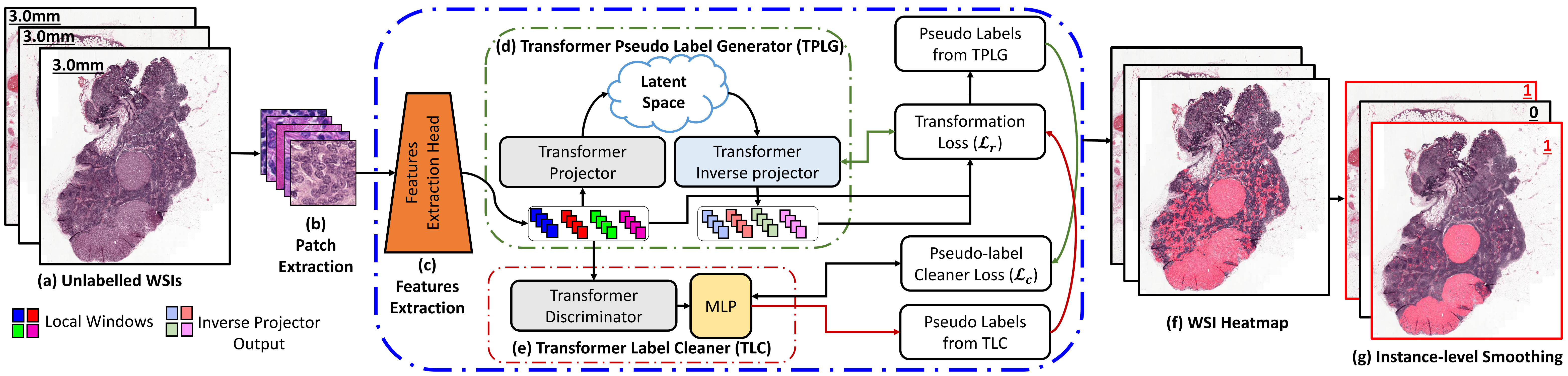}
\caption{System diagram of the proposed UMTL algorithm for WSI classification. 
(a) Shows the unlabeled WSIs, (b) instances of size $224 \times 224 \times 3$ pixels are extracted, (c) feature extraction head, (d) Transformer Pseudo-Label Generator (TPLG), (e)  Transformer pseudo-Label Cleaner (TLC), (f) predicted WSI map where red region shows the positive instances, (g) instance-level label smoothing and slide-level label prediction steps.
}
\label{fig3}
\vspace{-1.4\baselineskip}
\end{figure*}

In more detail, we propose a mutual learning framework based on 
transformer architecture that has recently demonstrated excellent performance in many computer vision applications \cite{vaswani2017attention, carion2020end, chen2021pre, khan2021transformers}. 
The proposed system encompasses a transformer pseudo-label generator that assigns positive/negative labels to patches based on the reconstruction error and a label-cleaning network. 
The first module consists of a transformer projector and an inverse projector module which are trained to minimize the transformation error between the original and the inverse-transformed feature vectors. 
The label-cleaning network is also a transformer model trained to clean the noisy pseudo-labels using a transformer label-cleaner. 
The cleaned labels are then used to improve the transformer pseudo-label generator in the next iteration using the discriminative learning mechanism as discussed before and shown in Fig. \ref{fig3}.
Both transformer pseudo-label generator and pseudo-label cleaner modules mutually learn from each other, improving each other iteratively for instance-level classification.
For improved WSI classification, a graph smoothing mechanism is proposed as a post-processing step to suppress isolated spatially sparse positive labels.

The proposed algorithm has been trained in an end-to-end fully unsupervised manner. 
It is evaluated on four publicly available WSI classification datasets, including CAMELYON-16 \cite{bejnordi2017diagnostic} for breast cancer, The Cancer Genome Atlas (TCGA) lung cancer, TCGA for renal cell carcinoma and TCGA breast cancer \cite{tomczak2015review}.
Rigorous experimental evaluations demonstrate the excellent performance of the proposed unsupervised algorithm for WSI classification.
We have also performed experiments using a weakly supervised variant of our proposed method. We observed an enhancement in the performance with  this supervision support. 
Finally, we fine-tuned our proposed unsupervised pre-trained model to perform downstream analysis tasks such as cancer subtypes classification. 
In this experiment, the proposed algorithm outperformed the existing State-Of-The-Art (SOTA) MIL-based methods.
We summarize our main contributions as follows:

\begin{enumerate}
    \item We propose a fully unsupervised mutual transformer learning algorithm for instance-level predictions for WSI classification. 
    To the best of our knowledge, it is the first rigorous attempt to tackle the WSI classification problem in a fully unsupervised manner.
    \item The proposed architecture consists of two modules including a transformer pseudo-label generator and transformer label-cleaner, with both modules learning mutually from each other and improving the performance for instance-level classification.
    \item The transformer pseudo-label generator is based on the novel idea of learning a latent space via discriminative learning such that disease-negative instances can be inverse transformed with small errors while disease-positive instances observe large transformation errors.
    \item We perform rigorous experimental evaluations on four different WSI classification datasets. Cancer subtype classification is also evaluated as a downstream analysis task with weak supervision. Our results demonstrate the excellent performance of the proposed algorithm compared to several SOTA methods.
\end{enumerate}

The rest of this work is organized as follows: Section \ref{sec:literaturereview} presents a literature review on WSI classification methods.
Section \ref{sec:proposed} describes our proposed methodology in detail. 
Section \ref{sec:results} presents the experimental evaluation while Section \ref{sec:conclusion} draws the conclusion and describes the future directions of the current work.

\section{Literature Review}
\label{sec:literaturereview}
Deep learning has advanced computational pathology applications, however, the evolution has been hampered by the need for large-scale manually annotated WSI datasets.
To address this problem, MIL-based weakly supervised methods have been proposed, thereby avoiding expensive and time-consuming pixel-wise annotations \cite{shao2021transmil, zhang2022dtfd, li2021dual,lu2021data}. 
It has been empirically observed that a fully supervised classifier trained on a small pixel-level manually annotated dataset may overfit while a weakly-supervised classifier trained on a larger WSI-level labeled dataset may generalize better \cite{campanella2019clinical}.  
In the literature, MIL-based weakly-supervised methods have recently obtained much popularity towards WSI classification \cite{srinidhi2021deep}.
These methods can be broadly categorized into local and global representation-based methods \cite{hou2016patch, kanavati2020weakly, ilse2018attention, lu2021data}.
In the local methods, the label of each tissue instance is independently estimated and all labels are aggregated to estimate the WSI-level labels by averaging or max-pooling operation. 
In the global methods, representations of all instances within a bag are aggregated to obtain a global bag representation which is then used for the WSI classification.

\noindent \textbf{Local Methods:} Hou \textit{et al.} proposed a patch-based CNN model to differentiate between different cancer sub-types \cite{hou2016patch}.
The patch-level classification results are aggregated by using a decision-based fusion model.
Kanavati \textit{et al.} proposed instance-level fully supervised and weakly supervised learning to predict lung cancer from WSIs \cite{kanavati2020weakly}.
Lerousseau \textit{et al.} proposed a weakly-supervised MIL method for tumor segmentation in WSIs using region-level annotations \cite{lerousseau2020weakly}.
Xu \textit{et al.} proposed instance-level labels prediction and WSI segmentation method using slide-level labels \cite{xu2019camel}.
In these methods, only a small number of instances in each WSI contributes to the training therefore a large number of WSIs are required.

\noindent \textbf{Global Methods:} Ilse \textit{et al.} proposed a neural network-based permutation-invariant aggregation operator to obtain global representation from histology images \cite{ilse2018attention}.
Lu \textit{et al.} proposed a clustering-based attention method to be applied to the MIL problem for improving WSI classification performance \cite{lu2021data}.
Sharma \textit{et al.} proposed an end-to-end network for clustering the WSI instances into different groups \cite{sharma2021cluster}.
From each group, a few instances are sampled for training and an attention method is used for WSI classification.
These methods assume the instance to be generated from an independent and identically distributed process however, the spatially adjacent instances within WSI are highly correlated with each other.
Therefore, Shao \textit{et al.} proposed transformer-based correlation, as well as both morphological and spatial information for WSI classification \cite{shao2021transmil}.
Several other MIL-based variants are proposed for improved performance in medical imaging \cite{wang2018revisiting, zhu2017deep, srinidhi2021deep}.
Although, global-methods are better than local methods, however, for highly imbalanced classification problems the information of rare classes may get lost within the majority class during the features aggregation process.

\noindent \textbf{Self-supervised Learning Methods}:
Self-supervised learning aims to produce rich feature representations using a formulated supervision by the data itself.
The learned representations are then employed to improve the performance of the downstream analysis tasks. 
These techniques can be broadly categorized into contrastive learning-based and pre-text-based methods.

The contrastive learning-based methods extract augmentation invariant information and instance discriminating features by pulling closer similar samples and pushing away dissimilar ones \cite{le2020contrastive}.
The pre-text-based methods include magnification prediction, stain channel prediction, cross-stain prediction, color reconstruction, and neighborhood image-related transformation. 
Several contrastive learning-based methods have been recently proposed in computational pathology.
Li \textit{et al.} proposed a self-supervised contrastive learning framework to extract good representations to be used in MIL methods \cite{li2021dual}.
Ciga \textit{et al.} proposed a self-supervised contrastive learning method on large-scale histopathology datasets across multiple organs with different types of stains and resolutions \cite{ciga2022self}.
The learned features are then used to train a linear classifier in a supervised manner for the downstream task.
Huang \textit{et al.} extracts patch features via self-supervised learning and aggregates these feature representations based on spatial information and correlation between different patches \cite{huang2021integration}.
These features are then used for survival analysis as a downstream task.
Li \textit{et al.}  also proposed a contrastive learning-based features extraction method using self-invariance, inter-invariance, and intra-invariance between WSI patches \cite{li2021sslp}.
The features are then used for a linear classifier for cancer subtypes classification.
Abbet \textit{et al.} proposed a self-supervised learning method that simultaneously learns the tissue region representation as well as the clustering metric \cite{abbet2020divide}.
The learned representations are then used to predict survival using colorectal cancer WSIs. 
Vu \textit{et al.} learned holistic WSI-level representation using a handcrafted framework based on deep CNN \cite{vu2023handcrafted}.
The learned representations are then utilized for distinct cancer subtypes classification as a downstream analysis task.
Their proposed handcrafted histological transformer (H2T) is reported to be faster an order of magnitude faster than the state-of-the-art transformers.
More self-supervised learning methods can be seen in \cite{wang2022transformer} and \cite{chen2022scaling}.

Although self-supervision can also be employed in our proposed framework to further improve performance, currently our method is different from the existing self-supervised learning methods. 
We do not propose any pre-text task neither we employed contrastive learning for unsupervised WSI classification.
In contrast to existing methods which learn features using self-supervision and then employ them in supervised downstream analysis tasks, we propose a fully unsupervised WSI classification algorithm.
Our proposed algorithm without using slide-level labels or region-level annotations learns to identify cancerous patches in a large repository of WSIs.
Similar to the existing self-supervised learning methods, we also extend our work for downstream analysis tasks using supervised and semi-supervised settings. 
To the best of our knowledge, no rigorous fully unsupervised WSI classification algorithm has been found in the literature.

\section{Proposed Methodology}
\label{sec:proposed}
The schematic illustration of our proposed algorithm dubbed as Unsupervised Mutual Transformer Learning (UMTL) for WSI classification is depicted in Fig. \ref{fig3}.
The UMTL consists of four main steps including feature extraction heads, Transformer Pseudo-Label Generator (TPLG), Transformer pseudo-Label Cleaner (TLC), and instance-level label smoothing for WSI classification.
We first formulate the problem and then we explain each step in detail.

\subsection{Problem Formulation}
In the unsupervised WSI classification problem context, we consider each WSI as a bag consisting of multiple instances (a.k.a patches). 
Specifically, let $W_{j} = \{p_{i,j}\}_{i=1}^{n}$ be the $j$-th WSI consisting of $n$ instances and $p_{i,j}\in \mathbb{R}^{m \times m \times 3}$ denotes the $i$-th instance, $1 \le j \le b$.
In unsupervised settings, neither the slide-level labels nor the region-level annotations are used for training.
Our main goal is to estimate the slide-level label $Y_{j}\in \{0, 1\}$ using instance-level pseudo labels $\ell_{i,j} \in \{0, 1\}$:

\begin{equation}
    Y_j=\begin{cases} 1 \text{ if } \sum_{i=1}^n \ell_{i,j} \ge \beta_{WSI}  \\
    0  \text{ otherwise },
    \end{cases}
\label{eqn1}
\end{equation}

\noindent where $\beta_{WSI}$ is the minimum number of disease positive instances for a WSI to be considered as positive-label.

\subsection{Feature Extraction Head}
Each instance $p_{i,j}$ is input to a feature extraction head consisting of five convolutional layers which are learned such that overall loss is minimized in an end-to-end manner.
The output of the feature extraction head is $f_{i,j}=F_{h}(p_{i,j})\in \mathbb{R}^{m \times m \times c}$ which preserves the input instance size except for the number of channels which are increased to $c\ge3$.
The learned features $f_{i,j}$ are re-arranged as a sequence of local windows $w_{i,j,k}\in \mathbb{R}^{a \times a \times c}$ considered as words, where $1\le k \le n_k$, $n_k=m^2/a^2$. 
We also employ learnable positional encoding $u_{i,j,k} \in \mathbb{R}^{a \times a \times c}$ for each local window $w_{i,j,k}$ \cite{dosovitskiy2020image,carion2020end}. 
A position-aware representation $g_{i,j,k}=u_{i,j,k} + w_{i,j,k}$ is then computed and used for further processing.
\subsection{Transformer Pseudo Label Generator (TPLG)}
Transformers have been found to be powerful frameworks for many tasks including image classification, object detection, and representation learning \cite{shamshad2022transformers,vaswani2017attention,chen2021pre, khan2021transformers, carion2020end}.
In this work, we employ a similar transformer architecture proposed by Vaswani \textit{et al.} \cite{vaswani2017attention}.
Instances are projected to latent space by using a transformer-based projector and then inverse-transformed to the original space using a transform inverse projector.
The transformation loss is then used to assign pseudo-labels to each instance of the WSI. 
\subsubsection{Transformer Projector}
Our transformer projector consists of a Multi-head Self Attention (MSA) layer followed by a Multi-layer Perceptron (MLP) containing two fully connected layers.
Each WSI instance is re-arranged as a sequence of position-aware word representation, $g_{i,j,k}$ which is input to the transformer projector.
The projector transforms it to a learnable latent space such that $q_{i,j,k}$ be the latent representation of $g_{i,j,k}$.
The input to the projector is $p_{0}=[g_{i,j,1}, g_{i,j,2},..., g_{i,j,n_{k}}]$ and the subsequent projection steps are formulated as follows:
\vspace{-0.5mm}
\begin{equation}
\begin{split}
   & q_{x}=k_{x}=v_{x}=\textbf{LN}(p_{x-1}),~\hat{p}_{x}=\textbf{MSA}(q_{x},k_{x},v_{x})+p_{x-1}, \\
    &p_{x}=\textbf{MLP}(\textbf{LN}( \hat{p}_{x}))+\hat{p}_{x}, \\
    &p_{L}=[q_{(i,j,1)}, q_{(i,j,2)},..., q_{(i,j,n_{k})}],
    \end{split}
\label{eqn2}
\end{equation}

\noindent where $x=1,2,...,L$ denotes the number of projector layers and $LN$ represents the Layer Normalization \cite{ba2016layer}.
\subsubsection{Transformer Inverse Projector} 
The inverse projector assumes an opposite role to that of the projector.
More specifically, the inverse projector learns an inverse mapping from the latent space to that of the original feature space.
Therefore, the architecture of the inverse projector is similar to that of the transformer projector consisting of two MSA layers followed by MLP.
The difference to that transformer projector is we employ an inverse projection embedding as an additional input to the inverse projector.
This inverse projection embedding $b_{i,j,k} \in \mathbb{R}^{a^{2} \times c}$ is learned to facilitate the inverse projection of features to the original space.
The computation of the transformer inverse projector is then formulated for the $x$-th layer where $1\le x \le L$ and $L$ are the total number of layers in the back-projector.
\begin{equation}
\begin{split}
&z_{0}=p_{L}, q_{x}=k_{x}=\textbf{LN}(z_{x-1})+b_{i,j,k}, v_{x}=\textbf{LN}(z_{x-1}),\\
&\hat{z}_{x}=\textbf{MSA}(q_{x},k_{x},v_{x})+Z_{x-1}, \hat{q}_{x}=\textbf{LN}\hat{z}_{x}+b_{i,j,k}, \\
&\hat{k}_{x}=\hat{v}_{x}=\textbf{LN}(z_{0}), \Tilde{z}_{x}=\textbf{MSA}(\hat{q}_{x}, \hat{k}_{x}, \hat{v}_{x})+\hat{z}_{x}, \\
&z_{x}=\textbf{MLP}(\textbf{LN}(\Tilde{z}_{x}))+\Tilde{z}_{x}.
\end{split}
\label{eqn3}
\end{equation}

\noindent The output of the $L$-th layer of the transformer inverse projector is $z_{L}=[\hat{g}_{i,j,1}, \hat{g}_{i,j,2},..., \hat{g}_{i,j,n_{k}}]$.
The transformation loss $\mathcal{L}^w_{1}(i,j,k)$ at window $(i,j,k)$ is defined as:

\begin{equation}
\begin{split}
&\mathcal{L}^w_{1}(i,j,k)=||g_{i,j,k}-\hat{g}_{i,j,k}||_{1},~\mathcal{L}^p_{1}(i,j)=\sum_{k=1}^{n_k}  \mathcal{L}^w_{1}(i,j,k), \\
&\mathcal{L}^{WSI}_{1}(j)= \sum_{i=1}^{n}  \mathcal{L}^P_{1}(i,j),~\mathcal{L}_{r}=\sum_{j=1}^{b_{t}} \mathcal{L}_1^{WSI}(j),
\end{split}
\label{eqn4}
\end{equation}

\noindent $\mathcal{L}^p_{1}(i,j)$ is the loss at instance-level, $\mathcal{L}^{WSI}_{1}(j)$ is the loss at the WSI-level, and $\mathcal{L}_{r}$ is the loss of overall training data having $b_{t}$ number of WSIs.
During the training of the transformer projector and inverse projector, $\mathcal{L}_{r}$ loss is minimized.
For the purpose of pseudo-label generation for the $i$-th instance in the $j$-th WSI, a simple threshold approach may be used as:

\begin{equation}
    \ell_{i,j}=\begin{cases} 1 \text{ if } \frac{\mathcal{L}^p_{1}(i,j)-\min_{Batch}(\mathcal{L}^p_{1}(i,j))}{\max_{Batch}(\mathcal{L}^p_{1}(i,j))} \ge \beta_{r}  \\
    0  \text{ otherwise },
    \end{cases}
\label{eqn5}
\end{equation}

\noindent where $\beta_{r}$ is an instance-level threshold computed using the training data as discussed in the ablation study (see Fig. \ref{fig5}).
In the following sub-sections, a pseudo-label cleaner is proposed to further refine the pseudo-labels generated by TPLG. 

\begin{figure}[t!]
\centering
\includegraphics[width=\linewidth]{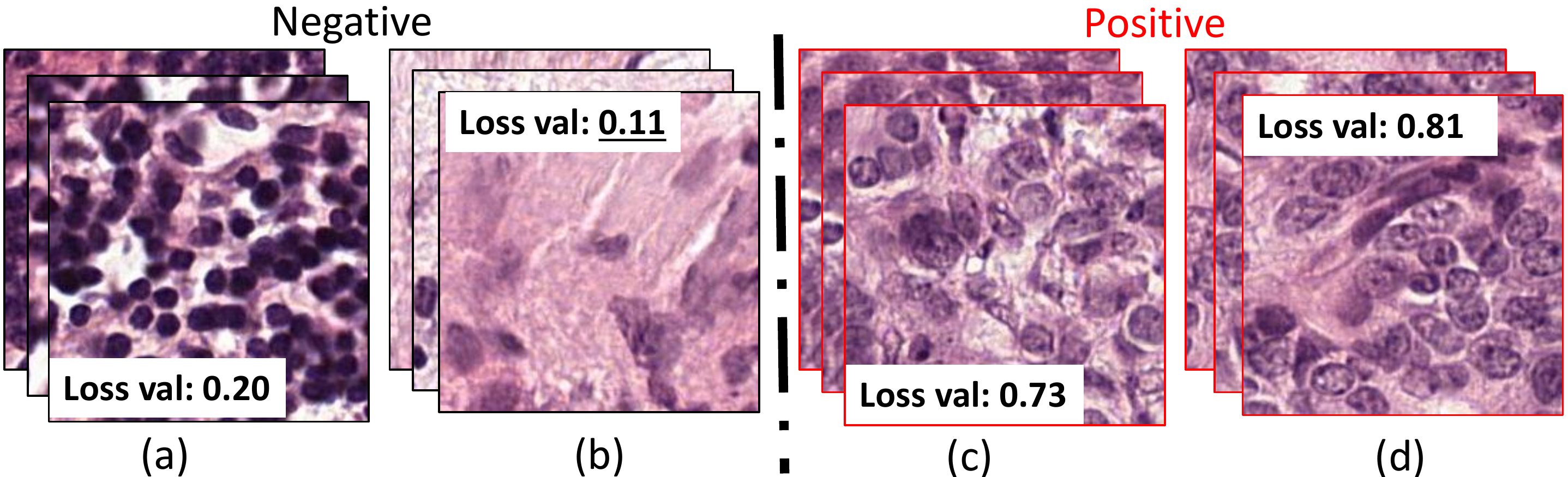}
\caption{Exemplar instances from positive and negative labels along with instance-level loss. 
(a) \& (b) Show instances of lymphocytes and stromal while (c) \& (d) show tumor instances. 
Transformation loss is low for negative-labeled instances and high for positive ones.}
\label{fig4}
\vspace{-1.3\baselineskip}
\end{figure}

\subsection{Transformer Pseudo Label Cleaner (TLC)}
In order to clean the noise in the pseudo-labels, we propose to train a Transformer-based pseudo-Label Cleaner (TLC) module.
The training of this module for classification task is performed in an end-to-end manner using the pseudo-labels obtained by (\textcolor{red}{5}).
Once, TLC is trained it is then used to generate new pseudo-labels based on the probabilities $\phi_{i,j}$ using the cross-entropy loss as:

\begin{equation}
\mathcal{L}_{c}=\frac{-1}{b_{t}}\sum_{j=1}^{b_{t}}\sum_{i=1}^{n}\ell_{i,j}\phi_{i,j}+(1-\ell_{i,j})\text{ln}(1-\phi_{i,j}),
\label{eqn6}
\end{equation}

\noindent The clean labels $\ell^{c}_{i,j}$ are predicted using:

\begin{equation}
    \ell^{c}_{i,j}=\begin{cases} 1 \text{ if } \phi_{i,j} \ge \beta_{c},  \\
    0  \text{ otherwise }, 
    \end{cases}
\label{eqn7}
\end{equation}

\noindent where $\beta_{c}$ is a threshold used to decide positive or negative label and it is estimated using the training data (see Fig. \ref{fig5}).
These labels $\ell^{c}_{i,j}$ will then be utilized for the training of the TPLG in the next iteration.
Both TPLG and TLC modules iteratively refine each other by mutual learning in an end-to-end manner.
As a result, the performance of the proposed UMLT improves over consecutive iterations.

\subsection{Discriminative Learning of TPLG}
In the second and onward iterations of TPLG, the labels from the TLC module are available for further training.
For the negative labels, the transformation loss is measured between the original and the inverse projected features.
However, for the positive labels, the transformation loss is measured between the inverse projected features and a fixed random Gaussian noise vector as shown in Fig. \ref{fig2}. 
Such an approach will result in increased transformation error for positive-labeled instances and decreased loss for the negative-labeled instances resulting in the improved discriminative ability of the UMTL algorithm (see Fig. \ref{fig4}).
In order to  make TPLG more discriminative, Eq. \textcolor{red}{4} is employed only for the negative-labeled instances while for the positive ones, the following formulation is used for the transformation loss minimization:

\begin{equation}
    \mathcal{L}^w_{1}(i,j,k)=||\tilde{g}_{f}-\hat{g}_{i,j,k}||_{1},
\label{eqn8}
\end{equation}

\noindent where $\tilde{g}_{f} \in \mathbb{R}^{a \times a \times c}$ is a random Gaussian noise having normal distribution $N(0,1)$.
We empirically observed that having a fixed noise matrix as a target better deludes TPLG than using a varying target for each positive instance.

\subsection{Instance Clustering}
\label{sec:clust}
The unsupervised training of the UMTL algorithm is under-constraint due to the lack of ground-truth labels.
In order to improve the performance of the UMTL, we propose an instance clustering-based pre-processing step to clean the training data by reducing tissue heterogeneity.

In most WSIs, the tumor region is relatively sparse while the normal region is more dominant.
To discriminate between the instances belonging to these regions, we employed a simple K-means clustering method.
The training data is grouped into $k_o$ clusters using the representations obtained from the features extraction head.
The $k_l$ larger clusters are considered normal instances and used for the training in the first iteration.
This pre-processing step does not completely separate the two types of instances however, it relatively cleans the input data for better training of the TPLG module in the first iteration.
In the later iterations, such pre-processing is not required because we start getting pseudo-labels from the proposed TLC which are then used for discriminative learning of TPLG.

\subsection{Instance Label Smoothing}
\label{sec3.4.3}
In order to predict the final label of WSIs, the instance-level labels are smoothed using graph convolutions \cite{kipf2016semi}.
A spatial graph $\mathcal{G}_{s}$ is constructed such that each instance is connected to its four spatial neighbors having adjacency matrix $A$.
The transformation loss $\mathcal{L}_{1}^{p}$ obtained from the trained TPLG is considered as the node attribute.
The node attribute vector $\ell_{s}$ is multiplied by graph adjacency matrix $A$ for attribute smoothing.
The $n$ such multiplications will propagate attributes to $n$ hop neighbors resulting in attribute smoothing.
Isolated attributes will be smoothed according to their neighbors.
The resulting attributes are given by $\hat{\ell_{s}}=\sigma(A^{n}\ell_{s})$, where $\sigma$ is a activation function.
Based on the connectivity of the positive-labeled instances overall label of the WSI is then predicted. 
A WSI is predicted as disease positive if the size of the positive-labeled connected components is larger than a $\beta_{WSI}$ threshold value.

\subsection{Weakly Supervised UMTL Algorithm}
\label{sec:subtype}
Most existing methods for WSI classification are trained in a weakly-supervised fashion. 
Therefore, we also incorporate weak supervision in our proposed unsupervised UMTL algorithm and dubbed it W-UMTL.
In the first setting, we train UMTL with weak supervision for cancer vs. normal WSI classification. 
For more details of this setting, please refer to the section \ref{sec:weak}. 

The second problem relates to the cancer subtype classification which requires further classification beyond just cancer vs. normal binary classification. 
For this purpose, we perform downstream analysis by first differentiating cancer vs. normal instances using the proposed UMTL algorithm trained in fully unsupervised settings. 
Then, only a TLC module is fine-tuned for cancer subtype classification of only positive instances using inherited WSI-level labels.
Therefore, we dub our downstream algorithm in this setting as Downstream UMTL (D-UMTL).
At test time, the normal vs. cancer instances are first differentiated using UMTL and then only positive instances are further classified for a particular cancer subtype using D-UMTL.
Cancer subtyping at the WSI level is performed using the same instance-level smoothing process as described in Sec. \ref{sec3.4.3}.
\vspace{-1.0\baselineskip}
\section {Experimental Evaluations}
\label{sec:results}
We compare the performance of the UMTL algorithm with its different variants and SOTA weakly supervised MIL-based methods on four different WSI classification datasets.
To validate the effectiveness of UMTL, we use different experimental protocols including fully unsupervised, limited weakly supervised, and training for downstream analysis tasks.
We have also performed ablation studies to demonstrate the contribution of each component of the proposed algorithm.
\vspace{-1.0\baselineskip}
\subsection{Datasets}
We have evaluated our proposed unsupervised WSI classification algorithm on four publicly available datasets including CAMELYON-16 \cite{bejnordi2017diagnostic} for breast cancer, TCGA for Lung Cancer (TCGA-LC), TCGA Renal Cell Carcinoma (TCGA-RCC), and TCGA BReast CAncer (TCGA-BRCA) for predicting HER status \cite{tomczak2015review}.
The details of each of these datasets are given in the below subsections.

\subsubsection{CAMELYON-16 Dataset}
It contains 400 WSIs with a split of 270/130 for training/testing purposes.
The training dataset consists of 159 normal slides or negative cases and 111 WSIs containing tumor regions of breast cancer metastasis considered as positive cases.
Tumor regions are annotated at pixel-level and labels at slide-level are assigned by an expert pathologist.
However, for the purpose of training in our fully unsupervised UMTL algorithm,  neither region-level annotations nor slide-level labels are used.
For testing purposes, slide-level labels are used to evaluate the performance of the compared methods.
The main challenge in this dataset is that the positive slides contain only small portions of the tumor.

\subsubsection {TCGA Lung Cancer Dataset}
TCGA-LC dataset consists of 1046 slides of two cancer subtypes including LUng Squamous cell Carcinoma (LUSC) \cite{cancer2012comprehensive} and LUng ADenocarcinoma (LUAD) \cite{cancer2014comprehensive} and 589 normal WSIs.
Compared to CAMELYON-16, tumor regions are significantly larger and only slide-level labels are available in this dataset.
We randomly split the 1635 WSIs into 80$\%$ and 20$\%$ training and testing split while ensuring patient-level separation.
On this dataset, two different types of experiments are performed.
Fully unsupervised WSI classification is performed for cancer vs normal using UMTL.
In downstream analysis tasks, LUSC vs LUAD classification is performed using Weakly-supervised UMTL (W-UMTL) with slide-level labels only.

We performed five-fold cross-validation experiments by randomly selecting the training and testing splits each time and average results are reported.
Within 1046 WSIs, this dataset contains 534 LUAD and 512 LUSC slides, respectively.
We randomly split the WSIs into 836 training slides and 210 testing slides for LUAD versus LUSC classification while ensuring patient-level separation.

\subsubsection{TCGA Renal Cell Carcinoma (RCC) Dataset}
This dataset contains 477 normal WSIs and 726 WSIs with three cancer subtypes including Kidney Renal Papillary Cell Carcinoma (KIRP) (218 WSIs) \cite{cancer2016comprehensive}, Kidney Renal Clear Cell Carcinoma (KIRC) (390 WSIs)\cite{cancer2013comprehensive}, and Kidney Chromophobe Renal Cell Carcinoma (KICH) (118 WSIs) \cite{davis2014somatic}.
Similar to the TCGA-LC, random 80$\%$ \& 20$\%$ training/testing splits are made while ensuring patient-level separation, and then 5-fold cross-validation experiments are performed.

Similar to TCGA-LC, experiments are performed in two different settings: fully unsupervised cancer vs normal using UMTL, and cancer subtype classification (KIRP vs. KIRC vs. KICH) as downstream task using W-UMTL with slide-level labels only.

\subsubsection {TCGA BReast CAncer (TCGA-BRCA) Dataset}
TCGA-BRCA dataset is used for the prediction of  Human Epidermal growth factor Receptor 2 (HER2) status which is a critical task in clinical practice for cancer treatment and prognostication \cite{cancer2012comprehensive}.
This dataset contains 608 WSIs with slide-level labels of HER2- status and 101 HER2+ status.
For training and validation, 80$\%$ data with patient-level separation is used while the remaining 20$\%$ is used for testing.  
We employed 5-fold cross-validation for comparison with other SOTA methods. 
On this dataset, first cancer vs. normal patch-level classification is performed using fully unsupervised UMTL. 
Then, only using the positive patches, HER2 +ve vs. -ve downstream classification is performed using W-UMTL. However, results are  only reported for cancer subtype classification because fully normal WSIs are unavailable in this dataset.

\subsection{Evaluation Metrics}
All experiments are evaluated using well-known measures including Accuracy (Acc), Area Under the Curve (AUC), and $F_{1}$ measures as reported by recent SOTA methods \cite{li2021dual, zhang2022dtfd, guan2022node, shao2021transmil}.
Since region-level annotations are also available in CAMELYON-16, therefore, we also performed lesion-based evaluation using Free-response Receiver Operating Characteristic (FROC) measure. 
It is defined as the average sensitivity at predefined six false positive rates: 1/4, 1/2, 1, 2, 4, and 8 FPs per WSI.

\subsection{Implementation Details}
For patch extraction from WSIs, we first employed the OTSU thresholding method to separate the tissue region from the background.
The tissue region is then divided into non-overlapping patches of size $224 \times 224$ at 20$\times$ magnification level.
In CAMELYON-16, the number of extracted patches is around 3.7 Million (M), in TCGA-LC $12.6$M, in TCGA-RCC $8.9$M, and in TCGA-BRCA $5.8$M.
In the pre-processing step (Sec. \ref{sec:clust}), the instances are clustered with $k_{o}=10$, and $k_{l}=3$ largest clusters  are retained in all experiments.

The overall architecture consists of features head and transformer layers.
Our features extraction head consists of one convolutional layer followed by two ResBlocks each consisting of two convolutional layers.
The first convolutional layer contains $3$ input channels, $64$ feature maps, and $3 \times 3$ size of kernel window.
The convolutional layers in each ResBlock contain $64$ input channels, $64$ output channels, and $5 \times 5$ kernel size.
Each transformer projector and inverse projector contain 12 layers.

We conducted our experiments on a DGX NVIDIA workstation with 256 GB of RAM and 4 Tesla V100 GPUs.
We trained both networks in an end-to-end manner using the Adam optimizer with 120 epochs.
The initial learning rate was set as $5e^{-5}$ with a batch size of 256.
Thresholds for TPLG and TLC are data-driven and found to be $\beta_{r}=0.50$ in Eq. (\ref{eqn5}), and $\beta_{c}=0.50$ in Eq. (\ref{eqn7}).
In Eq. \eqref{eqn1}, $\beta_{WSI}=10\%$  of the number of instances is used in all our experiments.
The ablation study of these values is discussed in the ablation study Section \ref{ref:ablation}.

\begin{table}[t!]
\caption{\textbf{Ablation (Abl) studies on the UMTL using CAMELYON-16 test-set.} 
Instance Clustering \textbf{IC} is a pre-processing step, Transformer Psuedo-Label Generator (\textbf{TPLG}), Auto-encoder Psuedo-Label Generator (\textbf{APLG}), Discriminative Learning (\textbf{DL}), MLP Label Cleaner (\textbf{MLC}), Transformer Label-Cleaner (\textbf{TLC}), and Instance-Label Smoothing (\textbf{ILS}) components are evaluated.}
\begin{center}
\makebox[\linewidth]{
\scalebox{0.80}{
\begin{tabu}{|[2pt]c|c|c|c|c|c|c|c|c|[2pt]}
\tabucline[2pt]{-}
Variant&IC&TPLG&DL&TLC&ILS&$F_{1}$&Acc&AUC\\\tabucline[2.0pt]{-}
UMTL&\checkmark&\checkmark&\checkmark&\checkmark&\checkmark&\textbf{0.751}&\textbf{0.832}&\textbf{0.844}\\\tabucline[0.5pt]{-}
UMTL$_{v1}$&&\checkmark&\checkmark&\checkmark&\checkmark&0.729&0.813&0.822\\\tabucline[0.5pt]{-}
UMTL$_{v2}$&\checkmark&\checkmark&&&\checkmark&0.712&0.791&0.803\\\tabucline[0.5pt]{-}
UMTL$_{v3}$&\checkmark&\checkmark&&\checkmark&\checkmark&0.733&0.810&0.822\\\tabucline[0.5pt]{-}
TLC$_{C}$&\checkmark&&&\checkmark&\checkmark&0.677&0.751&0.772\\\tabucline[0.5pt]{-}
UMTL$_{v4}$&\checkmark&\checkmark&\checkmark&MLC&\checkmark&0.728&0.813&0.831\\\tabucline[0.5pt]{-}
Auto-MLP&\checkmark&APLG&\checkmark&MLC&\checkmark&0.662&0.713&0.731\\\tabucline[0.5pt]{-}
UMTL$_v5$&\checkmark&\checkmark&\checkmark&\checkmark&&0.737&0.807&0.822\\\tabucline[2.5pt]{-}
\end{tabu}
}
}
\end{center}
\label{table1}
\end{table}

\begin{table*}[t!]
\caption{Performance of the proposed algorithm for cancer vs. normal WSI classification in two different settings including fully unsupervised algorithm UMTL and Weakly-supervised UMTL (W-UMTL) on three datasets. 
For UMTL, $0\%$ labels are used for both FROC and AUC.
For the W-UMTL variant, different percentages of WSIs labels are used and AUC is reported using the testing splits of each dataset. The lesion-based evaluation is also performed in CAMELYON-16 dataset in fully unsupervised manner and FROC is reported.}
\begin{center}
\makebox[\linewidth]{
\scalebox{0.75}{
\begin{tabu}{|c|c|c||c|c|c|c|c|c|c|c|c|c|}
\tabucline[2pt]{-}
\multirow{ 2}{*}{Datasets}&0$\%$&0$\%$&10$\%$&20$\%$&30$\%$&40$\%$&50$\%$&60$\%$&70$\%$&80$\%$&90$\%$&$100\%$\\
&FROC&AUC&AUC&AUC&AUC&AUC&AUC&AUC&AUC&AUC&AUC&AUC\\\tabucline[2.0pt]{-}
CAMELYON-16&0.388&0.844&0.801&0.833&0.867&0.901&0.922&0.941&0.949&0.951&0.961&0.966\\\tabucline[0.5pt]{-}
TCGA-LC&-&0.856&0.788&0.811&0.835&0.865&0.894&0.918&0.935&0.951&0.971&0.975\\\tabucline[0.5pt]{-}
TCGA-RCC&-&0.822&0.855&0.866&0.881&0.902&0.922&0.941&0.966&0.977&0.985&0.991\\\tabucline[2.5pt]{-}
\end{tabu}
}
}
\end{center}
\label{table2}
\end{table*}

\subsection{Unsupervised WSI Classification Results}
Cancer vs normal WSI classification is performed in a fully unsupervised manner using our proposed UMTL algorithm on three independent datasets including CAMELYON-16, TCGA-LC, and TCGA-RCC. 
No existing fully unsupervised methods could be found in the literature therefore, we have to make comparisons with weakly supervised methods where necessary.\\

\noindent \textbf{CAMELYON-16 dataset:} For this dataset, two experiments are performed in a fully unsupervised manner including lesion segmentation and WSI classification.

\begin{figure}[t!]
\centering
\includegraphics[width=3.5in]{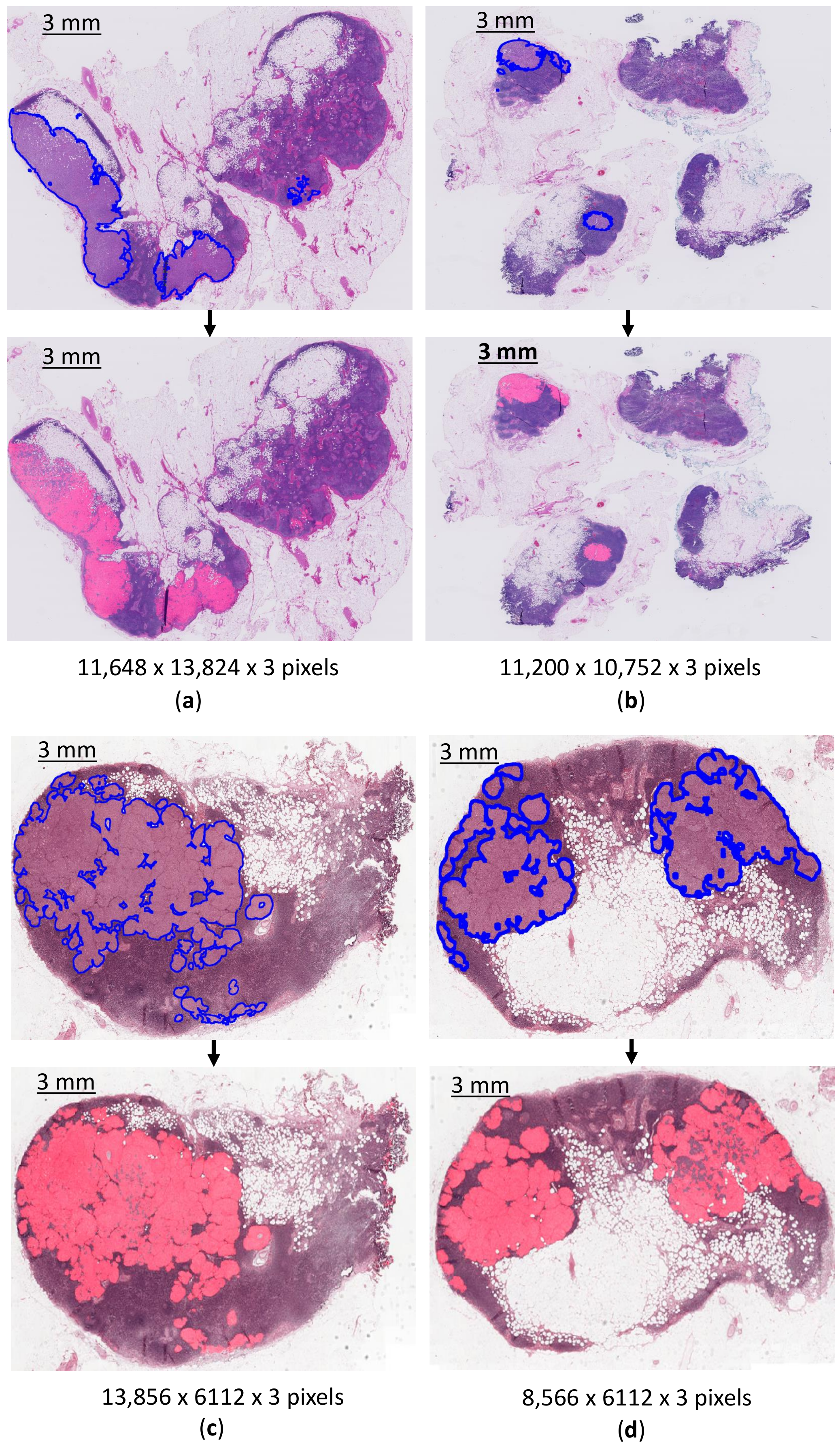}
\caption{Visualization of instance labels obtained by UMTL algorithm. (a)-(b) Show two different WSIs, (c)-(d) Show two subfields of the same WSI selected from CAMELYON-16 test set. Top row shows the ground-truth region-level tumor annotation with blue boundaries.
Bottom row shows the positive instances with pink color while the remaining region shows the negative instances.}
\label{fig6}
\end{figure}

For the case of lesion segmentation, using $0\%$ labels or annotations, cancerous lesions are segmented using our proposed UMTL algorithm.
In this experiment, we obtained 38.8$\%$ performance as reported in Table \ref{table2}.
Our performance is better than some existing weakly-supervised methods including Mean Pooling, Max-Pooling, and RNN-MIL, and comparable with classic AB-MIL as shown in Table \ref{table_CM}.
The unsupervised lesion segmentation obtained by UMTL algorithm is shown in Fig. \ref{fig6}.
A visual comparison with region-level ground-truth annotation reveals the effectiveness of the unsupervised lesion segmentation estimated by the proposed UMTL algorithm.

For unsupervised WSI classification results, we obtained 84.40$\%$ performance as shown in Table \ref{table2}.
Among the existing weakly-supervised methods, our proposed UMTL algorithm is comparable with PT-MTA, classic AB-MIL, and Max-Pooling while better than the Mean Pooling method (see Table \ref{table_CM}).\\

\noindent \textbf{TCGA-LC dataset:} For this dataset, fully unsupervised WSI classification is performed achieving 85.6$\%$ performance using the proposed UMTL algorithm as shown in Table \ref{table2}.
Our performance is better than the weakly supervised PT-MTA method.\\

\noindent \textbf{TCGA-RCC dataset:} For this dataset, in fully unsupervised settings our proposed UMTL algorithm obtained 82.20$\%$ AUC performance for WSI classification as shown in Table \ref{table2}.

\subsection {Ablation Studies and Analysis}
\label{ref:ablation}
Since there are no existing fully unsupervised WSI classification methods, therefore we use several variants of our proposed UMTL algorithm for detailed performance comparisons.
Some of these variants are designed by exclusion or inclusion of different components as mentioned in Table \ref{table1}.
Therefore, the performance variations reflect the relative contribution of each component while the UMTL has demonstrated the best performance compared to all variants. 
These experiments are performed using the CAMELYON-16 test set under fully unsupervised settings.

\subsubsection{Significance of Instance Clustering (IC) Pre-processing Step} 
In this experiment, the pre-processing Instance Clustering (IC) step (Sec. \ref{sec:clust}) is removed from the proposed UMTL algorithm to evaluate its significance.
The resulting algorithm is dubbed UMTL$_{v1}$.
The overall F$_1$ performance of UMTL$_{v1}$ is degraded by 2.20$\%$ compared to UMTL which shows the contribution of the pre-processing IC step.

\subsubsection{Performance of Transformer Pseudo-Label Generator (TPLG)}
In this experiment, only the TPLG module is employed while the Transformer-based Label Cleaner (TLC) module is excluded as a result, the DL step is also removed.
This version of the proposed UMTL algorithm is dubbed UMTL$_{v2}$.
Compared to UMTL, the performance of UMTL$_{v2}$ degraded by  3.90$\%$ which demonstrates that the TPLG in itself can also be used for fully unsupervised WSI classification.
However, the best combination is having both TPLG and TLC modules.

\subsubsection{Significance of Discriminative Learning (DL) Step}
In this experiment, the TPLG module is modified by the exclusion of the DL step.
This version of the proposed UMTL algorithm is dubbed UMTL$_{v3}$.
As a result, we only train the TPLG module using the reconstruction loss on both +ve and -ve instances.
Since the DL step enabled iterative refinement of TPLG therefore this refinement is also not possible in consecutive iterations. 
Compared to the proposed UMTL algorithm, the UMTL$_{v3}$ demonstrated 1.80$\%$ performance degradation.
Therefore, the iterative refinement of UMTL using DL step positively contributes to the performance of the overall learning algorithm.

\subsubsection{Clustering-based Pseudo Labels}
In this experiment, TPLG is removed, and the pseudo labels are generated by using IC such that the labels of the largest $k_{l}=3$ clusters are used as negative and the remaining cluster labels are used as positive. Label cleaning is then performed using TLC module.
This version is dubbed as TLC$_{C}$.
In this version, Instance Label Smoothing (ILS) component is employed similarly to the proposed UMTL algorithm.
Compared to the UMTL algorithm, the performance of TLC$_{C}$ is reduced by 7.40$\%$.
The significant reduction in performance may be attributed to the noise in the clustering-based pseudo labels.
Compared to that the pseudo labels generated by our proposed TPLG module have reduced noise and improved the overall performance.

\subsubsection{MLP Label Cleaner}
In this experiment, a simple MLP is used as a label cleaner module. 
This version is dubbed as UMTL$_{v4}$.
The input to the MLP is latent space features as shown in Fig. \ref{fig2} and MLP is trained using the cross-entropy loss function.
The performance of UMTL$_{v4}$ is 72.80$\%$ which is 2.30$\%$ less than the proposed UMTL algorithm demonstrating the relative importance of the transformer-based label cleaner. 

\subsubsection{Using Autoencoder and MLP}
In this experiment, we employed a simple Auto-encoder Psuedo-Label Generator (APLG) using ResNet50 instance-level features.
This version is dubbed Auto-MLP.
APLG consists of five fully connected layers [1024, 512, 256, 512, 1024] and MLP is used as a Label Cleaner (MLC).
The performance of Auto-MLP is 66.20$\%$ which is 8.90$\%$ less than the proposed UMTL algorithm showing the importance of transformer-based architecture both in TPLG and TLC.

\subsubsection{Significance of Instance Level Smoothing (ILS)}
In this experiment, the ILS is removed from the proposed UMTL algorithm as described in Sec. \ref{sec3.4.3}.
Instead of ILS, Eq. (\ref{eqn1}) is used for WSI-level classification with $\beta_{WSI}=10\%$.
This version is dubbed UMTL$_{v5}$.
Compared to UMTL, the performance of UMTL$_{v5}$ degraded by 1.40$\%$ in $F_{1}$ score and a 2.20$\%$ in AUC.
The reduction in performance demonstrates the significance of the ILS step.

\subsection{Ablation on Parameters Tuning}
\subsubsection{Selection of TPLG Threshold}
\begin{figure}[t!]
\centering
\includegraphics[width=\linewidth]{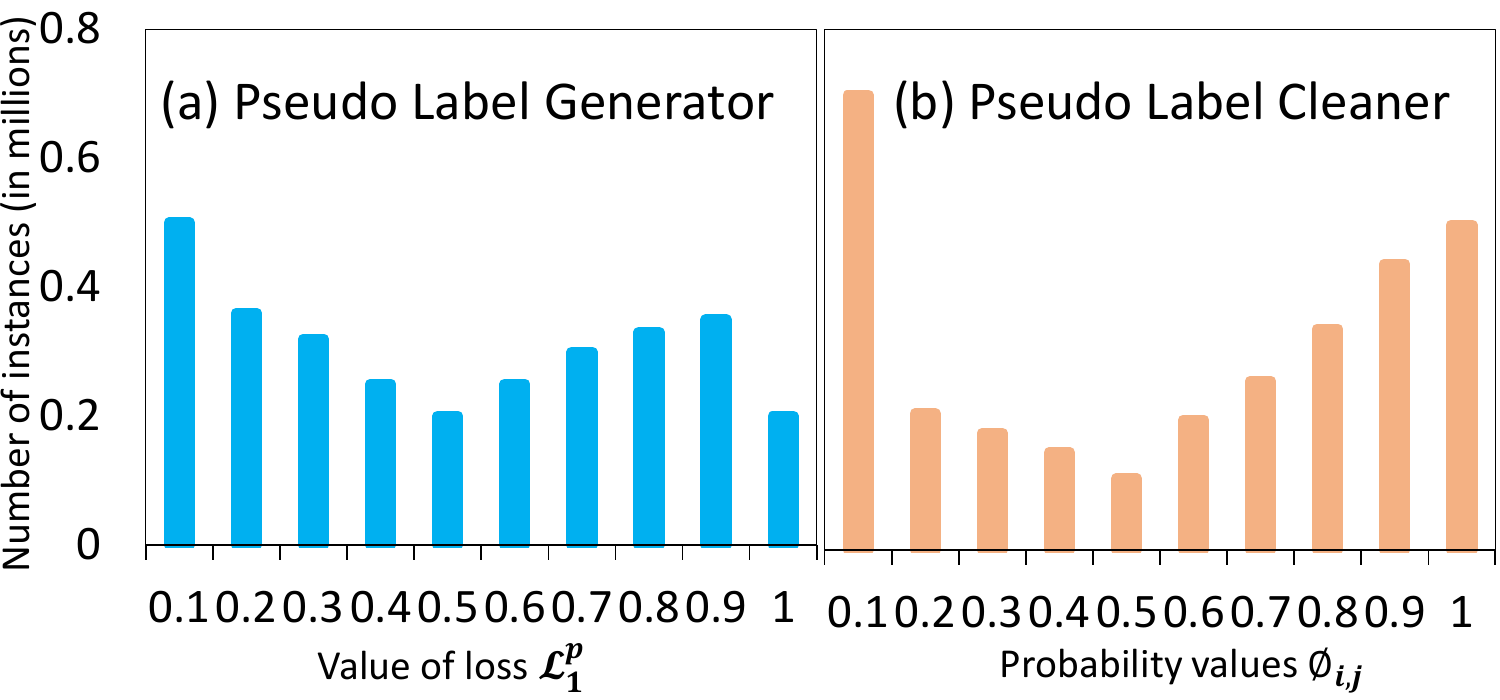}
\vspace{-1.5\baselineskip}
\caption{(a) Distribution of transformation loss and (b) classification probabilities for CAMELYON-16 training split. 
The value of $\beta_{r}$ and $\beta_{c}$ in Eqs. (\ref{eqn5})  and (\ref{eqn7}) is set to be 0.5.}
\label{fig5}
\vspace{-1.5\baselineskip}
\end{figure}

In TPLG module, a threshold on the transformation loss is required to decide whether an instance is positive or negative.
For this purpose, a threshold $\beta_{r}$ is introduced in Eq. (\ref{eqn5}).
To empirically select the value of $\beta_{r}$, the distribution of transformation loss is plotted over the training data as shown in Fig. \ref{fig5} (a).
The transformation loss is scaled from 0 to 1 by dividing by the maximum loss on any instance.
It is observed that instances with close to 0 errors are negative while those having close to 1 are positive.
In Fig. \ref{fig5} (a) we observed a dip in the percentage of instances at 0.50 transformation error.
Therefore, we select $\beta_{r}=0.50$.

\subsubsection{Selection of  TLC Threshold}
In the TLC module, a probability is generated for an instance to be positive or negative.
The distribution of this probability over the training dataset is plotted in Fig. \ref{fig5} (b).
We observed a dip in the distribution at the probability of 0.50 therefore, we select $\beta_{c}=0.50$ in Eq. (\ref{eqn7}).
Moreover, in this plot, we also observe a higher percentage of instances toward 0 and 1 probabilities compared to the transformation loss plot.
It demonstrates the performance of the label-cleaner for pushing the tumor instances towards the probability of 1 and normal instances towards the probability of 0.

\subsubsection{Ablation on the Number of Clusters Parameter}
For Instance Clustering (IC) pre-processing step, input instance data is grouped into $k_{o}$ clusters, and $k_{l}$ larger clusters are considered negative while the remaining clusters are considered as positive.
In the first experiment, $k_{l}=3$ is fixed and $k_{o}$ is varied as 5,10,15,20, and 25.
The best AUC is observed at $k_{o}=10$ as shown in Table \ref{table_cluster}.

In the second experiment, $k_{o}=10$ is fixed and $k_{l}$ is varied as 1,1-3,1-5,1-7, and 1-9.
The best AUC is observed at $k_{l}=1-3$ as shown in Table \ref{table_cluster}.
This means that the three largest clusters out of a total of 10 clusters produced the best performance.

\begin{table}[t!]
\vspace{-1.5\baselineskip}
\caption{AUC on CAMELYON-16  by varying $k_{o}$ \& $k_{l}$ after \textbf{1st} Epoch.}
\begin{center}
\makebox[\linewidth]{
\scalebox{0.70}{
\begin{tabu}{|c|c|c|c|c|c|}
\tabucline[2.0pt]{-}
Fixed&$k_{o}=5$&$k_{o}=10$&$k_{o}=15$&$k_{o}=20$&$k_{o}=25$\\\tabucline[0.5pt]{-}
$k_{l}=3$&0.771&\textbf{0.798}&0.797&0.784&0.781\\\tabucline[2.0pt]{-}
Fixed&$k_{l}=1$&$k_{l}=1~to~3$&$k_{l}=1~to~5$&$k_{l}=1~to~7$&$k_{l}=1~to~9$\\\tabucline[0.5pt]{-}
$k_{o}=10$&0.751&\textbf{0.798}&0.781&0.772&0.766\\\tabucline[2.0pt]{-}
\end{tabu}
}}
\end{center}
\label{table_cluster}
\end{table}

\vspace{-1.0\baselineskip}
\subsection{Comparison with Weakly Supervised Methods}
\label{sec:weak}
The main focus of the current work is fully unsupervised WSI classification, however, currently, no such methods have been found in the literature.
The nearest methods we observed are weakly supervised Multiple Instance Learning (MIL)-based WSI classification methods including  Mean-Pooling and Max-Pooling as used by SOTA \cite{zhang2022dtfd}, RNN-MIL \cite{campanella2019clinical}, classic AB-MIL \cite{ilse2018attention}, DS-MIL \cite{li2021dual}, CLAM-SB \cite{lu2021data}, CLAM-MB \cite{lu2021data}, PT-MTA \cite{li2019patch}, Trans-MIL \cite{shao2021transmil}, DTFD-MIL \cite{zhang2022dtfd}, MS-ABMIL \cite{hashimoto2020multi}, C2C \cite{sharma2021cluster}, ZoomMIL \cite{thandiackal2022differentiable}, and NAGCN \cite{guan2022node}.

For a fair comparison, we trained the proposed UMTL algorithm with supervision and dubbed it W-UMTL.
For this purpose, the number of labeled WSIs in training data is gradually increased from 10$\%$ to 100$\%$ as shown in Table \ref{table2}.
Since instance-level labels are not available, therefore, we make each instance inherit the label from its parent WSI
A label-cleaning mechanism is then employed based on the TPLG loss which is used as a pre-trained model.
Instances with a  loss $>$0.50 from normal WSIs and those having a loss $<0.50$ from positive WSIs are discarded to clean the inherited labels for an improved training process.
The remaining instances are used in the end-to-end training of the TLC module.
We reported the performance of the proposed weakly-supervised learning algorithm for cancer vs. normal WSI classification on three datasets including CAMELYON-16, TCGA-LC, and TCGA-RCC in Table \ref{table2}.
The performance of the proposed algorithm improves with increasing the level of supervision while the best performance is observed with 100$\%$ weak supervision.

Table \ref{table_CM} shows the weakly-supervised WSI classification results on the CAMELYON-16 test set and compared with existing SOTA methods.
We report W-UMTL results with 100$\%$ slide-level labels for cancer vs. normal WSI classification.
For the weakly-supervised setting, we obtained an AUC of 96.60$\%$ which is better than the SOTA methods including TrasnMIL and DTFD-MIL. 
The weakly-supervised lesion-based evaluation resulted in 47.60$\%$ FROC which is better than all compared SOTA methods (Table \ref{table_CM}).

Cancer vs. normal WSI classification experiments is also performed on TCGA-LC and TCGA-RCC datasets by varying the slide-level labels from 10$\%$ to 100$\%$ (Table \ref{table2}).
Unfortunately, on these datasets, such a classification has not been found in the literature therefore, we are not able to compare these results with any existing SOTA methods.

\begin{table}[t!]
\caption{Performance comparison of the proposed W-UMTL algorithm with SOTA methods on the testing splits of CAMELYON-16. CAMELYON-16 test-set is evaluated for cancer vs. normal WSI classification.}
\vspace{-1.0\baselineskip}
\begin{center}
\makebox[\linewidth]{
\scalebox{0.98}{
\begin{tabu}{|c|c|c|c|c|}
\tabucline[2pt]{-}
Methods&$F_{1}$&Acc&AUC&FROC\\\tabucline[2.0pt]{-}
Mean Pooling&0.355&0.626&0.528&0.116\\\tabucline[0.5pt]{-}
Max-Pooling&0.754&0.826&0.854&0.331\\\tabucline[0.5pt]{-}
RNN-MIL \cite{campanella2019clinical}&0.798&0.844&0.875&0.304\\\tabucline[0.5pt]{-}
Classic AB-MIL\cite{ilse2018attention}&0.780&0.845&0.854&0.405\\\tabucline[0.5pt]{-}
DS-MIL\cite{li2021dual}&0.815&0.899&0.916&\textcolor{blue}{\textbf{0.437}}\\\tabucline[0.5pt]{-}
CLAM-SB \cite{lu2021data}&0.775&0.837&0.871&-\\\tabucline[0.5pt]{-}
CLAM-MB \cite{lu2021data}&0.774&0.823&0.878&-\\\tabucline[0.5pt]{-}
PT-MTA \cite{li2019patch}&-&0.827&0.845&-\\\tabucline[0.5pt]{-}
Trans-MIL \cite{shao2021transmil}&0.797&0.883&0.930&-\\\tabucline[0.5pt]{-}
DTFD-MIL\cite{zhang2022dtfd}&\textcolor{blue}{\textbf{0.882}}&\textcolor{blue}{\textbf{0.908}}&\textcolor{blue}{\textbf{0.946}}&-\\\tabucline[0.5pt]{-}
MS-ABMIL \cite{hashimoto2020multi}&-&0.876&0.887&-\\\tabucline[0.5pt]{-}
Proposed W-UMTL&\textcolor{red}{\textbf{0.895}}&\textcolor{red}{\textbf{0.911}}&\textcolor{red}{\textbf{0.966}}&\textcolor{red}{\textbf{0.476}}\\\tabucline[2.0pt]{-}
\end{tabu}
}
}
\end{center}
\label{table_CM}
\vspace{-1.5\baselineskip}
\end{table}

\begin{table}[t!]
\caption{Performance comparison of the proposed D-UMTL algorithm with SOTA methods for cancer subtypes classification on TCGA-LC (LUAD vs. LUSC) and TCGA-RCC (KIRCH vs. KIRP vs. KIRC) datasets.} 
\begin{center}
\makebox[\linewidth]{
\scalebox{0.95}{
\begin{tabu}{c|c|c|c|c|c|}
\tabucline[2pt]{2-6}
 &\multicolumn{ 3}{|c|[2.0pt]}{TCGA-LC}&\multicolumn{ 2}{|c|[2.0pt]}{TCGA-RCC}\\\tabucline[2.0pt]{-}
Methods&$F_{1}$&Acc&AUC&Acc&AUC\\\tabucline[2.5pt]{-}
Mean Pooling&0.809&0.833&0.901&0.905&0.978\\\tabucline[0.5pt]{-}
Max-Pooling&0.833&0.846&0.901&0.937&0.987\\\tabucline[0.5pt]{-}
RNN-MIL \cite{campanella2019clinical}&0.831&0.845&0.894&-&-\\\tabucline[0.5pt]{-}
Classic AB-MIL\cite{ilse2018attention}&0.866&0.869&0.941&0.893&0.970\\\tabucline[0.5pt]{-}
DS-MIL\cite{li2021dual}&0.876&0.888&0.939&0.929&0.984\\\tabucline[0.5pt]{-}
CLAM-SB \cite{lu2021data}&0.864&0.875&0.944&0.881&0.972\\\tabucline[0.5pt]{-}
CLAM-MB \cite{lu2021data}&0.874&0.878&0.949&0.896&0.979\\\tabucline[0.5pt]{-}
C2C \cite{sharma2021cluster}&-&0.873&0.938&0.919&0.987\\\tabucline[0.5pt]{-}
PT-MTA \cite{li2019patch}&-&0.737&0.829&0.905&0.970\\\tabucline[0.5pt]{-}
Trans-MIL \cite{shao2021transmil}&0.876&0.883&0.960&0.946&0.988\\\tabucline[0.5pt]{-}
DTFD-MIL\cite{zhang2022dtfd}&\textcolor{blue}{\textbf{0.891}}&0.894&\textcolor{blue}{\textbf{0.961}}&-&-\\\tabucline[0.5pt]{-}
MS-ABMIL \cite{hashimoto2020multi}&-&0.900&0.955&-&-\\\tabucline[0.5pt]{-}
NAGCN \cite{guan2022node}&-&\textcolor{blue}{\textbf{0.902}}&0.952&\textcolor{blue}{\textbf{0.954}}&\textcolor{red}{\textbf{0.992}}\\\tabucline[0.5pt]{-}
HIPT \cite{chen2022scaling}&-&0.895&0.952&0.923&0.980\\\tabucline[0.5pt]{-}
Prop. D-UMTL&\textcolor{red}{\textbf{0.911}}&\textcolor{red}{\textbf{0.933}}&\textcolor{red}{\textbf{0.976}}&\textcolor{red}{\textbf{0.972}}&\textcolor{blue}{\textbf{0.991}}\\\tabucline[2.5pt]{-}
\end{tabu}
}
}
\end{center}
\label{table5}
\end{table}

\vspace{-1.0\baselineskip}
\subsection{Evaluations on Downstream Analysis Tasks}
In order to compare the proposed UMTL algorithm with existing weakly-supervised methods for downstream analysis tasks we extend our method by the inclusion of weak supervision and dubbed it D-UMTL.
More details can be found in Sec. \ref{sec:subtype}.
We compared the proposed D-UMTL algorithm with weakly-supervised methods as well as self-supervised methods.
Both of these categories of methods use weak supervision for downstream analysis tasks.
\subsubsection{Comparison with Weakly-Supervised Methods}
These comparisons are performed on three distinct datasets including TCGA-LC, TCGA-RCC, and HER2.

\noindent \textbf{Experiment on TCGA-LC dataset} is performed for LUAD vs. LUSC cancer subtypes classification task and the results are reported in Table \ref{table5}.
The proposed D-UMTL algorithm with weak supervision obtained 97.60$\%$ AUC score outperforming all SOTA methods.
The closest competitor is DTFD-MIL obtaining 96.10$\%$ AUC.

Similar to TCGA-LC dataset, an \textbf{experiment on TCGA-RCC} is performed for KIRCH vs. KIRP vs. KIRC cancer sub-types WSI classification.
The results are reported in Table \ref{table5}.
The proposed D-UMTL algorithm with weak supervision obtained 97.20$\%$ Acc and 88.10$\%$ $F_{1}$ score, outperforming existing SOTA methods while obtaining comparable AUC (99.10$\%$).
The closest competitor is NAGCN obtaining 95.40$\%$ Acc and 99.20$\%$ AUC.

Table \ref{table4} shows the results of predicting \textbf{HER2 status} (either HER2+ or HER2-) on the TCGA-BRCA dataset. 
For the weakly-supervised setting, D-UMTL obtained an AUC of 79.10$\%$ better than the SOTA approaches including the recently proposed SlideGraph \cite{lu2022slidegraph+} method.
These results show the effectiveness of our transformer-based architecture for downstream analysis tasks using weak supervision.

\begin{table}[t!]
\caption{Performance of the proposed D-UMTL algorithm for HER2 status prediction on TCGA-BRCA. 
The AUC is reported using the test split.}
\begin{center}
\makebox[\linewidth]{
\scalebox{0.98}{
\begin{tabu}{|c|c|}
\tabucline[2pt]{-}
Methods&AUC\\\tabucline[2.0pt]{-}
RNN-MIL \cite{campanella2019clinical}&0.670\\\tabucline[0.5pt]{-}
Kather \textit{et al.}\cite{kather2020pan}&0.620\\\tabucline[0.5pt]{-}
Kather \textit{et al.} \cite{kather2019deep}&0.680\\\tabucline[0.5pt]{-}
Rawat \textit{et al.} \cite{rawat2020deep}&0.710\\\tabucline[0.5pt]{-}
CLAM \cite{lu2021data}&0.650\\\tabucline[0.5pt]{-}
SlideGraph \cite{lu2022slidegraph+}&\textcolor{blue}{\textbf{0.750}}\\\tabucline[0.5pt]{-}
Proposed D-UMTL&\textcolor{red}{\textbf{0.791}}\\\tabucline[0.5pt]{-}
\end{tabu}
}
}
\end{center}
\label{table4}
\end{table}

\subsubsection{Comparison with Self-Supervised Learning Methods}
In self-supervised learning-based methods, first a data representation is learned in unsupervised manners without using labels then a classifier is trained using those representations in weakly-supervised manners for downstream analysis task.
For fair comparison, we also employ weak supervision only for downstream analysis task.
Therefore, our proposed algorithm is dubbed as D-UMTL for cancer subtypes classification.
Comparisons are performed on two datasets including TCGA-LC and TCGA-RCC and compared with three very recent self-supervised learning-based methods including HIPT \cite{chen2022scaling}, H2T \cite{vu2023handcrafted} and SRCL \cite{wang2022transformer} as shown in Table \ref{table_self}.
For LUAD vs. LUSC subtypes classification in TCGA-LC dataset, the proposed D-UMTL algorithm obtained best performance of 97.60$\%$ while for  KIRCH vs. KIRP vs. KIRC  sybtypes classification in TCGA-RCC dataset, D-UMTL performance is comparable with H2T and SRCL methods.
It should be noted that self-supervised learning can also be used to improve our proposed algorithm's performance. 

\begin{table}[t!]
\caption{Performance comparison of the proposed D-UMTL algorithm with self-supervised learning methods on two different datasets. The AUC is reported using the test split.}
\begin{center}
\makebox[\linewidth]{
\scalebox{0.90}{
\begin{tabu}{|c|c|c|c|c|}
\tabucline[2pt]{-}
Datasets&H2T \cite{vu2023handcrafted}&HIPT \cite{chen2022scaling}&SRCL \cite{wang2022transformer}& Prop. D-UMTL\\\tabucline[2.0pt]{-}
TCGA-LC&0.802&0.952&\textcolor{blue}{\textbf{0.973}}&\textcolor{red}{\textbf{0.976}}\\\tabucline[0.5pt]{-}
TCGA-RCC&\textcolor{red}{\textbf{0.993}}&0.980&\textcolor{blue}{\textbf{0.991}}&\textcolor{blue}{\textbf{0.991}}\\\tabucline[0.5pt]{-}
\end{tabu}
}
}
\end{center}
\label{table_self}
\end{table}
\vspace{-1.0\baselineskip}
\section{Conclusion \& Future Work}
\label{sec:conclusion}
In this work, a fully unsupervised WSI classification algorithm is proposed using a Transformer Pseudo Label Generator (TPLG) and Transformer Label Cleaner (TLC).
In TPLG, instances are projected to a latent space and then inverse-projected to the original space using a projector and inverse projector.
Based on the transformation error, instances are assigned pseudo labels of being normal vs. cancerous.
These pseudo labels are then cleaned using a label-cleaning mechanism employed by TLC.
Both components mutually learn from each other for obtaining better labels in an iterative manner.
Based on the cleaned labels estimated by TLC, a discriminative learning mechanism is employed in the TPLG module so that the transformation error increases for the positive instances and decreases for the negative instances.
Experiments are performed in fully unsupervised as well as weakly supervised settings for cancer vs. normal WSI classification on four different datasets.
For downstream analysis, cancer subtype classification is performed using weak supervision for TLC finetuning.
The proposed algorithm has demonstrated excellent performance compared to SOTA methods.
As a future direction, investigating clinical tasks such as survival prediction using the proposed algorithm may be performed.


\bibliographystyle{IEEEtranS}
\bibliography{tmi}

\end{document}